\documentclass[runningheads]{llncs}

\usepackage{eccv}

\usepackage{eccvabbrv}

\usepackage{graphicx}
\usepackage{booktabs}
\usepackage{fontawesome}
\usepackage{wrapfig,lipsum,booktabs}

\usepackage[misc]{ifsym}
\usepackage[accsupp]{axessibility}  

\usepackage[dvipsnames]{xcolor}

\usepackage{tabu}                     
\usepackage{multirow}                 
\usepackage{multicol}                 
\usepackage{float}                    
\usepackage{makecell}                 
\usepackage{booktabs}                 
\usepackage[ruled, linesnumbered]{algorithm2e}  
\usepackage{colortbl}
\usepackage{graphicx}
\usepackage{svg}
\usepackage[framemethod=TikZ]{mdframed}
\newcommand{\tablestyle}[2]{\setlength{\tabcolsep}{#1}\renewcommand{\arraystretch}{#2}\centering\footnotesize}
\usepackage{tabularx}
\usepackage{array}
\definecolor{gray}{HTML}{efefef}

\usepackage{hyperref}

\usepackage{orcidlink}
\newcommand\blfootnote[1]{%
  \begingroup
  \renewcommand\thefootnote{}\footnote{#1}%
  \addtocounter{footnote}{-1}%
  \endgroup
}

\begin{document}

\title{Part2Object: Hierarchical Unsupervised 3D Instance Segmentation} 

\titlerunning{Part2Object}

\author{Cheng Shi$^*$ \and
Yulin Zhang$^*$ \and
Bin Yang \and
Jiajin Tang \and \\
Yuexin Ma \and
Sibei Yang\textsuperscript{\Letter}
}

\authorrunning{Cheng et al.}

\institute{School of Information Science and Technology\\
ShanghaiTech University\\
\email{\{shicheng2022,yangsb\}@shanghaitech.edu.cn}\blfootnote{$*$ Equal contribution. \Letter~Corresponding author.}\\ 
 }

\maketitle

\begin{abstract}
  Unsupervised 3D instance segmentation aims to segment objects from a 3D point cloud without any annotations. 
  Existing methods face the challenge of either too loose or too tight clustering, leading to under-segmentation or over-segmentation. 
To address this issue, we propose Part2Object, hierarchical clustering with object guidance. Part2Object employs multi-layer clustering from points to object parts and objects, allowing objects to manifest at any layer. Additionally, it extracts and utilizes 3D objectness priors from temporally consecutive 2D RGB frames to guide the clustering process.
Moreover, we propose Hi-Mask3D to support hierarchical 3D object part and instance segmentation. 
By training Hi-Mask3D on the objects and object parts extracted from Part2Object, we achieve consistent and superior performance compared to state-of-the-art models in various settings, including unsupervised instance segmentation, data-efficient fine-tuning, and cross-dataset generalization. 
Code is release at \url{https://github.com/ChengShiest/Part2Object}.
\end{abstract}

\section{Introduction}
\label{sec:intro}

3D instance segmentation, parsing semantic compositions from complex point cloud scenes into distinct objects, is fundamental and crucial in real-world applications, such as mixed reality~\cite{suo2023mixsim}, autonomous navigation~\cite{suomela2023benchmarking, an2023etpnav}, planning~\cite{huang2023diffusion}, and manipulation~\cite{geng2023gapartnet}. In recent years, 3D instance segmentation has seen substantial advancements~\cite{qi2017pointnet++, wang2019dynamic, rethage2018fully, occuseg, vu2022softgroup++, Schult23mask3d}, improving both the accuracy and efficiency of object localization and recognition. Nevertheless, it often demands large-scale human annotations for fully-supervised training, resulting in costly and manpower-intensive efforts. While a few studies~\cite{nunes2022unsupervised, song2022ogc} have initiated to address 3D instance segmentation without human annotations, their reliance on 3D scene flows from sequential point clouds restricts their applicability to single point cloud scenarios, such as indoor scenes in ScanNet~\cite{dai2017scannet}.

In this paper, we study unsupervised 3D instance segmentation, aiming to segment class-agnostic instances from \textcolor{black}{indoor} 3D scenes without relying on any human labels. To address this problem, \textcolor{black}{two straightforward strategies have been explored recently}: 1) to apply traditional clustering~\cite{felzenszwalb2004efficient, mcinnes2017accelerated_hdbscan} or graph-cut methods~\cite{zhang2023freepoint} to group points into objects based on their RGB-D data, such as coordinates, colors, normal vectors, and self-supervised pretraining features~\cite{rozenberszki2023unscene3d}. 
2) To project 2D instance segmentation results from 2D RGB frames onto the point cloud using the corresponding camera pose. Typically, 2D RGB frames are captured concurrently with indoor point cloud data~\cite{scannet200,dai2017scannet}, while 2D instance masks are extracted using 2D unsupervised instance segmentation advancements~\cite{wang2022freesolo, wang2023cut, wang2022tokencut}. 
However, both strategies can only work in straightforward scenes with few salient objects, as they cannot tackle the fundamental challenges inherent in unsupervised 3D instance segmentation, as outlined below. 

\begin{figure}[t]
\centering
\caption{\textbf{Motivation of our hierarchical clustering.} Single-level clustering results in a trade-off between under-segmentation for certain objects and over-segmentation for others. In contrast, our hierarchical clustering allows for gathering and identifying objects at varying levels of clustering granularity.}
\vspace{-3mm}
\includegraphics[width=0.98\textwidth]{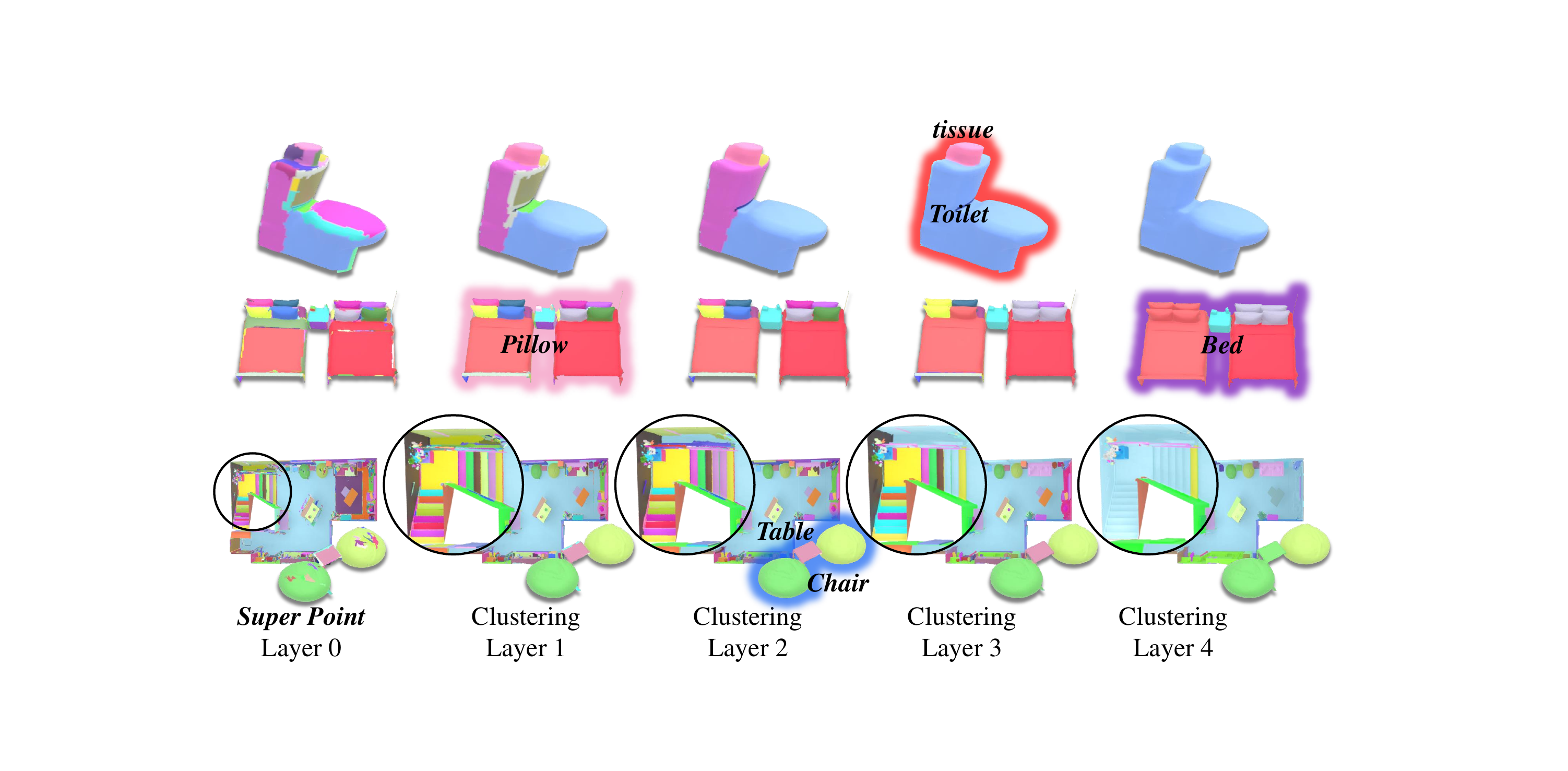}
\label{fig:intr1}
\end{figure}

In complex scenes, \textbf{\textit{achieving consistent segmentation granularity for different objects within a single clustering and graph-cut result is challenging}}, given the significant variations in geometric shape, color, and size among 3D objects, as well as the diverse compositions of these objects within scenes.
This entails a trade-off between under-segmentation for certain objects and over-segmentation for others, making it almost impossible to attain satisfactory segmentation granularity for all of them. 
As shown in Fig~\ref{fig:intr1}, tighter grouping can accurately segment 
objects with simple geometric shapes and textures, such as the pillow on the bed (\textcolor{black}{see layer one}). However, it can also result in the over-segmentation of larger and more complex objects like the toilet in the same layer (layer one), and vice versa. 
\textit{\textbf{In addition, there is a lack of an effective strategy to identify whether a point cluster or group represents a 3D instance.}} For example, the toilet back (see layer two) forms a well-grouped cluster, but it does not constitute a meaningful object. The 3D geometric features are fundamentally semantically insufficient to identify objects. 
Recently, it has been discovered that self-supervised vision transformer (ViT) features contain information about object segmentations in 2D images~\cite{caron2021emerging_dino,wang2023cut}. Building on this discovery, 3D unsupervised instance segmentation approaches~\cite{rozenberszki2023unscene3d} project the 2D self-supervised ViT features (or their inferred object segmentations) from 2D RGB frames onto 3D point clouds, thereby enhancing the perception of objectness for 3D point features. 
However, a 3D point typically corresponds to multiple pixels in different 2D RGB frames, and this many-to-one mapping is fragile, easily disrupted by noise features or imprecise masks on any frame, resulting in an inability to distinguish one object from others or the segmentation of objects into fragmented segments, as illustrated in Fig~\ref{fig:dino}\textcolor{red}{(b)} and \textcolor{red}{(c)}. In addition, despite enhanced point features, discerning nearby objects or those sharing similar semantics remains challenging due to the inherently greater complexity of 3D scenes compared to 2D images.
To tackle these challenges, we introduce a simple yet effective principle, \textcolor{black}{``Gather \& Aim''}, for the automatic discovery and segmentation of 3D objects. 
\textbf{\textit{For the gathering}}, inspired by the observation that tighter or looser groupings can respectively include certain objects within a scene, we propose allowing the gathering of potential objects at varying levels of grouping granularity, rather than being restricted to a single level. 
Specifically, we perform hierarchical clustering in the 3D scene, gradually grouping points into larger clusters, as illustrated in Fig~\ref{fig:intr1}. Therefore, instead of striving to improve single-layer clustering results, ours is expected that as long as each potential instance appears in one of the layers, \textcolor{black}{we will be able to lock onto the segmentation mask of that instance.} This mitigates excessive reliance on any single-level clustering results, thereby enhancing the clustering's adaptability to significant variations among objects and scenes. 
\textbf{\textit{For the aiming}}, we propose to target the 3D objectness priors, a set of candidate objects, from temporally consecutive 2D RGB frames. We leverage both the temporal consistency across frames and the prior knowledge of unsupervised object discovery within frames to extract the 3D objectness priors.  
Note that we jointly estimate 2D objects from multiple frames first and then approximate object-level 3D bounding boxes, departing from the point-level projection used in previous methods~\cite{rozenberszki2023unscene3d}, to address the fragility issue of the many-to-one mapping from pixels to points. 
It avoids adhesive or scattered 3D fragments caused by inconsistencies and occlusions over 2D RGB frames. 
Finally, the 3D bounding boxes serve as indicators to identify the 3D instance masks within the multi-level clusters. \textcolor{black}{Our ``Gather \& Aim''} strategy enables leverage of 2D semantic priors for 3D object identification while harnessing 3D geometric priors for precise instance segmentation.

Furthermore, the taxonomy of how object parts are composed into an object is a strong yet disregarded prior information for unsupervised 3D instance segmentation. For example, the taxonomy based on the parts of a chair, including the back, arm, seat, and leg, can assist in addressing highly challenging scenarios, such as distinguishing two closely positioned and semantically similar chairs as two separate instances. Thanks to our hierarchical clustering, we can not only identify objects but also trace their constituent parts (see the toilet and its parts in layers three and two, respectively). 
Therefore, we extract both objects and object parts from our hierarchical clustering and name our clustering algorithm Part2Object.  
Expanding on this, we enhance the 3D instance segmentation framework, Mask3D~\cite{Schult23mask3d}, to support hierarchical 3D part and instance segmentation. Our improved model, named Hi-Mask3D, takes the object and object parts from Part2Object as pseudo-labels for learning 3D instance segmentation by incorporating explicit interactions between object parts and objects. 


To evaluate the effectiveness of our Part2Object clustering and Hi-Mask3D model, we conduct experiments on the challenging and cluttered indoor environments~\cite{scannet200,dai2017scannet,s3dis} in three settings: 1) direct evaluation on unsupervised instance segmentation, 2) application to data-efficient fine-tuning, and 3) cross-dataset generalization, all showing significant performance improvement. In summary, our contributions are multi-fold: 

\begin{itemize}
\setlength{\itemsep}{0pt}
\setlength{\parsep}{0pt}
\setlength{\parskip}{0pt}
\item We propose two key insights for unsupervised 3D instance segmentation: 1) Employing hierarchical clustering enables the gathering of objects at varying levels of clustering granularity. 2) Leveraging 3D objectness priors from temporally consecutive 2D frames as guidance, while harnessing 3D geometric priors to clustering on the point cloud for precise instance segmentation. 
\item Based on our insights, we propose an innovative hierarchical clustering approach, Part2Object. It progressively groups points into object parts and objects while extracting and leveraging 3D objectness priors to guide the clustering process. Our Part2Object significantly outperforms the state-of-the-art training-free unsupervised 3D instance segmentation methods by 16.8\% mAP@50 on the ScanNet dataset.
\item We propose Hi-Mask3D, an extension of 3D instance segmentation to support hierarchical unsupervised 3D part and instance segmentation. Experiments demonstrate that Hi-Mask3D consistently and significantly outperforms state-of-the-art models in all the settings. 
\end{itemize}

\section{Related Work}
\label{sec:related_work}

\noindent \textbf{Unsupervised 3D instance segmentation.}
3D instance segmentation is an essential task in 3D scene understanding, which aims to locate and recognize different objects in a 3D point cloud. However, the cost and labor intensity of 3D scene-level instance annotation underscore the significance of exploring challenging unsupervised instance segmentation methods.
Early works utilize raw geometric information, including coordinates, colors, and normal vectors, to perform traditional clustering~\cite{felzenszwalb2004efficient, mcinnes2017accelerated_hdbscan,dbscan} for segmentation. Recently, inspired by the new paradigm in 2D unsupervised instance segmentation, some works~\cite{rozenberszki2023unscene3d, zhang2023freepoint} introduce pseudo labels and self-training to unsupervised 3D instance segmentation. 
Specifically, these methods use 2D or 3D self-supervised models to extract point cloud features, applying graph-cut algorithms to generate pseudo-labels for training and prediction. 
As illustrated earlier, the clustering method encounters a trade-off between under-segmentation and over-segmentation. Relying on a single graph-cut algorithm, as recent works do, falls short of achieving consistent granularity in complex indoor environments.

\noindent \textbf{Transfer 2D foundation models into 3D.}
The rapid advancements in 2D vision have given rise to powerful foundation models~\cite{kirillov2023segment,caron2021emerging_dino, amir2021deep_dino_dense,li2022languagedriven, ghiasi2021open, shi2023edadet,shi2023logoprompt,shi2024the,tang2023contrastive,dai2024curriculum}, proving beneficial across a variety of visual tasks. In contrast to the 2D domain, 3D vision encounters challenges with data scarcity and training limitations, hindering the development of foundation models. Consequently, several works in the 3D turn to leveraging 2D foundation models to address 3D problems. 
In 3D instance segmentation, some works~\cite{yang2023sam3d, chen2023towards} utilize 2D vision foundation models to generate pseudo-labels for 2D images, aiding in the training and prediction of 3D models through 2D to 3D projection algorithms. Other works~\cite{peng2023openscene,liu2023segment} employ features rather than outputs obtained from 2D foundation models, enhancing the feature extraction capabilities of 3D models through pixel-point alignment. However, relying solely on 2D features or 2D pseudo-labels poses challenges in addressing cluttered indoor scenes with stacked objects. 


\noindent \textbf{Supervised point cloud segmentation.}
Supervised point cloud segmentation can be divided into instance-level and part-level segmentation.
The former has been significant breakthroughs in recent years\cite{qi2017pointnet++, rethage2018fully, wang2018sgpn, hu2020randla, occuseg, chen2021hierarchical, hou20193d, vu2022softgroup++, wu2019pointconv, hui2022graphcut, Schult23mask3d, sun2022spformer, vu2022softgroup, wang2019dynamic, kolodiazhnyi2023topdown}.
However, these approaches demand a substantial amount of annotated data and focus on learning object-level attributes, neglecting the understanding of parts and their contributions to the composition of objects.
In the realm of 3D part segmentation, various setups, including supervised~\cite{qi2017pointnet++, zhang2021point, liu2020self}, weakly supervised~\cite{chibane2022box2mask, chen2019bae, wang2022ikea} instance segmentation and semantic segmentation, as well as open-world semantic segmentation~\cite{peng2023openscene,huang2023openins3d}, have been extensively explored. While these approaches have demonstrated promising result, their primary emphasis lies in determining how to divide an object into parts. 
To integrate instance-level and part-level perspectives, we perform hierarchical clustering with the prior knowledge of how parts compose objects, unifying instance-level object identification and part-level constituent tracing.

\section{Methodology}

\begin{figure*}[t]
\centering
\caption{\textbf{Overview of our Part2Object hierarchical clustering and Hi-Mask3D instance segmentation framework.} Part2Object extracts 3D objectness priors from consecutive 2D RGB frames and uses them to guide hierarchical clustering from points to object parts and objects. Hi-Mask3D utilizes objects and parts identified by Part2Object as pseudo-labels, learning for improving instance segmentation through the utilization of object parts.} 
\includegraphics[width=1\textwidth]{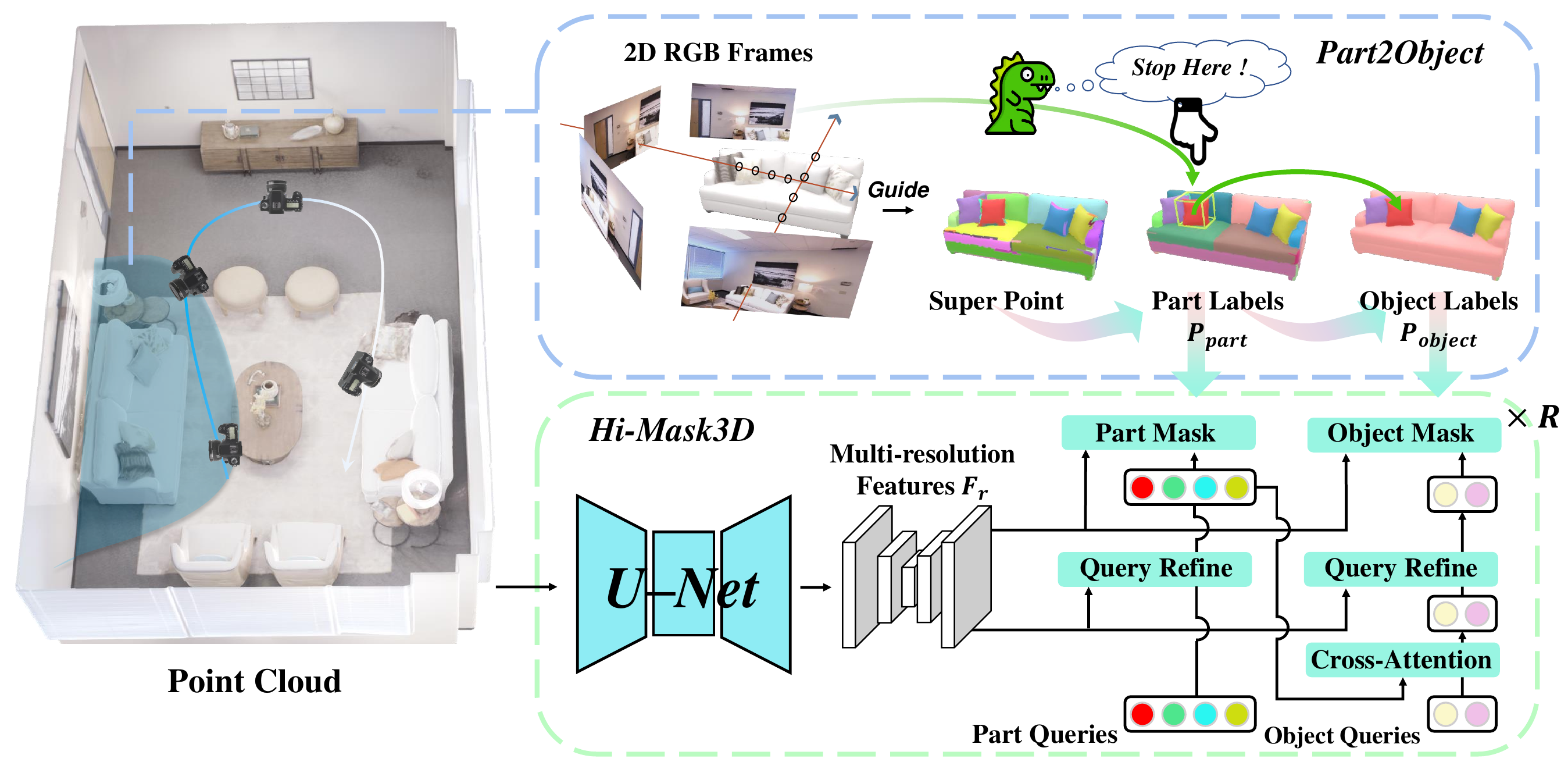}
\label{fig:pipeline}
\end{figure*}

\noindent\textbf{Problem definition.} Unsupervised 3D instance segmentation~\cite{rozenberszki2023unscene3d, zhang2023growsp} is the task of generating class-agnostic masks for all foreground objects in a 3D point cloud scene solely based on raw RGB-D data, without relying on any object detection or instance segmentation annotations. The raw RGB-D data~\cite{dai2017scannet, scannet200} consists of the point cloud $P \in \mathbb{R}^{N \times 3}$ and corresponding 2D RGB frames $\{I_i\}_{i=1}^M$, where each image frame $I_m \in \mathbb{R}^{H \times W \times 3}$. With raw RGB-D data, 3D instance segmentation task aims to output all object masks \textcolor{black}{$P_{\text{object}}$} in the scene.

\noindent\textbf{Method overview.} As shown in Fig~\ref{fig:pipeline}, our unsupervised framework has two major components: 1) the hierarchical clustering algorithm (Sec~\ref{subsec:clustering} and Sec~\ref{subsec:discovery}), Part2Object, which progressively group points into larger clusters. 
During clustering, aided by \textcolor{black}{objectness priors} from 2D frames, we can effectively obtain both object-level and part-level segmentation masks, which serve as pseudo-labels for learning unsupervised instance segmentation. 
2) 
The end-to-end 3D instance segmentation model Hi-Mask3D (Sec~\ref{subsec:mask3d}) extends the original Mask3D model~\cite{Schult23mask3d} by explicitly predicting segmentation results of object parts and leveraging them to aid instance segmentation. 
By self-training on the pseudo-labels extracted from the first-stage Part2Object, Hi-Mask3D can effectively predict 3D instance segments.

\subsection{Hierarchical Clustering on 3D Point Cloud}
\label{subsec:clustering}
\begin{figure*}[t]
\centering
\caption{\textbf{Robustness of our cluster features and 3D objectness priors.} (a) Visualization of the first 3 PCA components and our computed weights of points in Equ~\ref{equ:update}. (b) Visualization of 2D DINO features' PCA components, 3D points' normal vectors, and our 3D object priors. The comparison between (c) the projection-first-then-grouping pipeline and (d) our grouping-first-then-projection pipeline. 
} 
\includegraphics[width=1\textwidth]{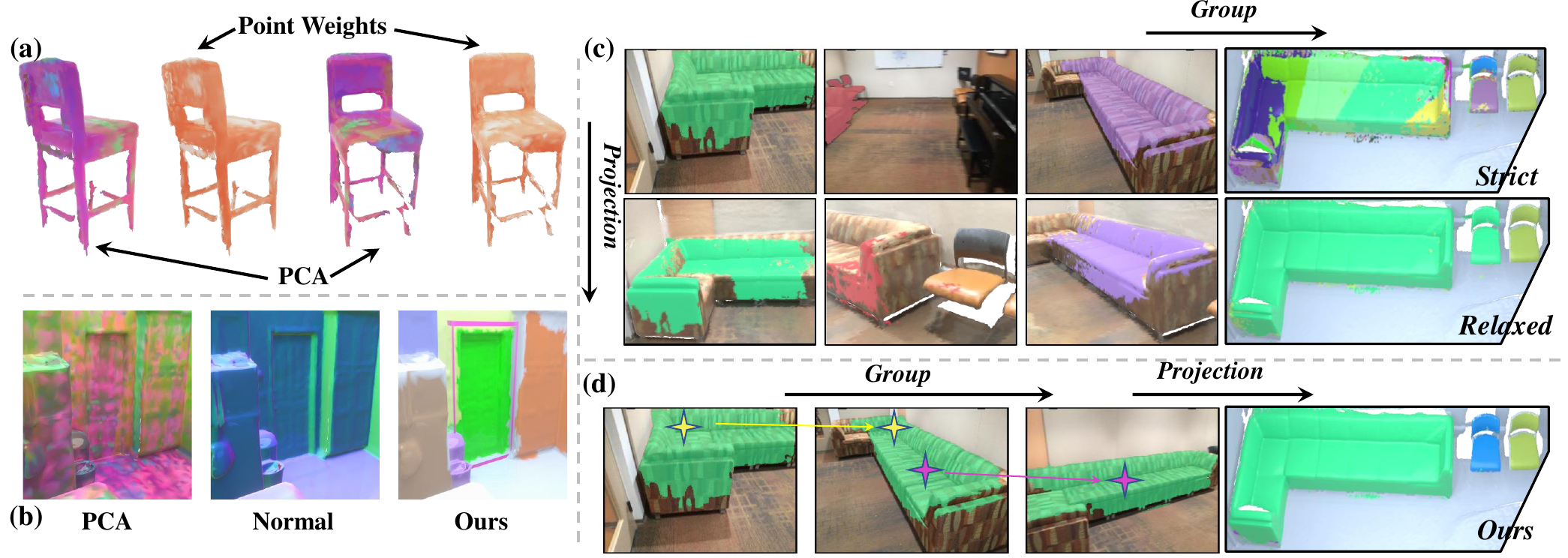}
\label{fig:dino}
\end{figure*}

In this section, we introduce the hierarchical clustering process of our Part2Object, which starts by grouping initial points into initial clusters and then progressively merging them into larger clusters. 
We will sequentially cover: 1) the feature representation of points, 2) the initialization of clusters from points, and 3) the hierarchical clustering process for merging clusters. 
\noindent\textbf{Point Feature Representation.}
Given a 3D surface point cloud $P$, we denote its norm and color as $P^{\text{Norm}}$ and $P^{\text{RGB}}$, respectively. 
\textcolor{black}{The 3D feature $\boldsymbol{f}_i^{\text{3D}}$ of point $p_i \in P$ comprises $p_i$, $p_i^{\text{Norm}}$, and $p_i^{\text{RGB}}$, representing the location, color, and geometry information, respectively. }
Additionally, we define point cloud 2D semantics feature $\boldsymbol{f}_i$ by projecting 2D features~\cite{caron2021emerging_dino} of RGB frames into 3D space using the corresponding camera pose. Here, \textcolor{black}{2D features}
are extracted using the 2D self-supervised method DINO~\cite{caron2021emerging_dino}, without rely on any 2D instance annotations. 
\noindent\textbf{Cluster Initialization and Feature Representation.} 
Given the vast number of points in the point cloud data~\cite{adams1994seeded}, we initially apply the VCCS algorithm~\cite{papon2013voxel} to generate super-points on the point cloud, forming the first-layer clusters $\{c_i^0\}_{i=1}^{N_0}$ as follows: 
\begin{equation}
\begin{aligned}
    & \{c_i^0\}_{i=1}^{N_0} = \text{VCCS}(P, P^{\text{Norm}}, P^{\text{RGB}}), \\
\end{aligned}
\end{equation}
where VCCS algorithm takes points' coordinates $P$, norms $P^{\text{Norm}}$, and colors $P^{\text{RGB}}$ as input and group original $N$ points into $N_0$ super-points as $\{c_i^0\}_{i=1}^{N_0}$. 
We consider each super-point $c_i^0$ as a cluster, consisting of multiple points that are closely positioned and share similar color and norm characteristics. 
Then, we define the feature representation for the cluster based on the features of the points within the cluster. 
Instead of directly averaging point features within the cluster, we compute a weight for each point and perform a weighted sum of features. These weights are determined by the similarity between each point's feature and the average feature of points within the cluster. This approach helps mitigate the influence of noisy points in the cluster, leading to more robust representations. Specifically, for the cluster $c_i^0$, its cluster feature $\boldsymbol{f}_i^0$ is computed as follows, 
\begin{equation}
\label{equ:update}
\begin{aligned}
\boldsymbol{f}_i^{0} &= \sum_{\substack{p_j \in c_i^0}} \frac{\mathrm{sim}(\boldsymbol{f}_{j}, \bar{\boldsymbol{f}_i^0})}{\sum_{\substack{p_j \in c_j^0}}\mathrm{sim}(\boldsymbol{f}_{j}, \bar{\boldsymbol{f}_i^0})} \boldsymbol{f}_{j}, \phantom{1} \texttt{where} \phantom{1} \bar{\boldsymbol{f}_i^0} &= \sum_{\substack{p_j \in c_i^0}} \frac{1}{|c_i^0|} \boldsymbol{f}_{j}.
\end{aligned}
\end{equation}
The $p_j \in c_i^0$ denotes each point $p_j$ within the cluster $c_i^0$, and $\boldsymbol{f}_{j}$ is the feature of point $p_j$. The $\bar{\boldsymbol{f}_i^0}$ is the average feature of points within $c_i^0$, and sim($\cdot,\cdot$) denotes the computation of cosine similarity. As depicted in the example in Fig~\ref{fig:dino}\textcolor{red}{(a)}, our cluster feature $\boldsymbol{f}_i^{0}$ demonstrates greater robustness compared to $\bar{\boldsymbol{f}_i^0}$.  
For simplicity, we abbreviate the operations in Equ~\ref{equ:update} as function \texttt{FU}($\cdot$), which computes the cluster feature for the inputted cluster based on point features, \eg, $\boldsymbol{f}_i^{0} = \texttt{FU}(c_i^0)$. 

\noindent\textbf{Hierarchical Clustering with One Stop Criteria.} 
Next, we group and merge the first-layer clusters $\{c_i^0\}_{i=1}^{N_0}$ to the next-layer clusters $\{c_i^1\}_{i=1}^{N_1}$ based on their features \{$ \boldsymbol{f}_i^{0}\}_{i=0}^{N_0}$ and spatial coordinates, progressively and iteratively forming higher-hierarchical clusters $\{c_i^t\}_{i=1}^{N_t}$, where $t$ represents the $t$-th layer in hierarchical clustering. 
In each single clustering layer, our clustering principle is that two clusters can only be merged when they are semantically similar and spatially adjacent: 1) the feature similarity between them ranks among the top $K$ similarities between any pair of clusters. 2) The closest points between two clusters are adjacent. 
Specifically, when considering two clusters in the $t$-th layer, $c_i^t$ and $c_j^t$, along with their features $\boldsymbol{f}_i^{t}$ and $\boldsymbol{f}_j^{t}$, the merging process to form the next-layer cluster $c^{t+1}_k$ and feature $\boldsymbol{f}^{t+1}_k$ is as follows:
\begin{equation}
\label{equ:cluster_metric}
\begin{aligned}
     c^{t+1}_k = c_i^t \cup c_j^t \ \phantom{1} \texttt{if} \phantom{1} \text{rank}(&\text{sim}(\boldsymbol{f}_i^{t}, \boldsymbol{f}_j^{t})) \leq K \phantom{1} \texttt{and } \text{dist}(c_i^t, c_j^t) \leq T, \\
     \boldsymbol{f}^{t+1}_k &= \texttt{FU}(c_i^t \cup c_j^t), \\
\end{aligned}
\end{equation}
where rank($\cdot$) denotes the ranking of feature similarity among pairwise cluster similarities in this clustering layer, where a higher rank signifies a higher similarity, and dist($\cdot$, $\cdot$) computes the Euclidean distance between closest points of the two clusters. The $K$ and $T$ represent the threshold values for ranking and distance, respectively. 
However, we observe that solely using the merging metric in Equ~\ref{equ:cluster_metric} can result in the incorrect merging of parts from different objects in the early stages of clustering (\ie, shallow layers of clustering). This occurs because clusters at the shallow layers are typically scattered fragments with highly local features, which hinder them from perceiving the entire object. To track this issue, we first extract 3D objectness priors, represented by a set of 3D bounding boxes of potential objects $B^{\text{3D}}$, 
detailed in Sec~\ref{subsec:discovery}. Then, we design a stopping criteria based on the \textcolor{black}{3D objectness priors} to prevent clusters belonging to different objects from being merged.
\textbf{\emph{Criteria}}: Stop merging clusters belonging to different objects!
Specifically, for any pair of clusters $c_i^t$ and $c_j^t$ that meet the merging metric, we extra assess their spatial relationship relative to each 3D object $b^{\text{3D}}_k \in B^{\text{3D}}$. If clusters $c_i^t$ and $c_j^t$ are respectively inside and outside the object $b^{\text{3D}}_k$, then reject their merging because they do not belong to the same object.

\noindent\textbf{Collect Objects and Parts from Clustering.} 
By employing hierarchical clustering with the stopping criteria, we identify clusters that stop merging with others as 3D objects, denoted as $\hat{P}_{\text{object}}$. 
Additionally, in hierarchical clustering, objects with complex geometric are typically merged from scattered fragments into meaningful parts, and then into complete objects. As shown in Fig~\ref{fig:intr1}, scattered ``toilet'' fragments are combined into object parts such as the toilet back and seat \textcolor{black}{(see layer two)}, which are then formed into the object ``toilet'' \textcolor{black}{(layer three)}. 
Therefore, we can trace back from objects $\hat{P}_{\text{object}}$ to identify the clusters from the previous level that form them, considering these clusters as potential object parts, \textcolor{black}{denoted as $\hat{P}_{\text{part}}$.} 
Both objects $\hat{P}_{\text{object}}$ and object parts $\hat{P}_{\text{part}}$ identified in hierarchical clustering serve as pseudo-labels for training our instance segmentation model, Hi-Mask3D~(Sec~\ref{subsec:mask3d}).

\subsection{3D Objectness Priors From 2D Frames}
\label{subsec:discovery}
In this section, we introduce our grouping-first-then-projection pipeline to extract \textcolor{black}{3D objectness priors} $B^{\text{3D}}$ from 2D RGB frames~\cite{dai2017scannet}, where $B^{\text{3D}}$ is the set of 3D bounding boxes for potential objects in the scene. 

\noindent\textbf{Grouping-first-then-projection Pipeline.}
To extract 3D objectness priors, a simple strategy involves projection-first-then-grouping: projecting 2D unsupervised instance segmentation masks~\cite{wang2023cut,shi2000normalized_cut} from 2D frames onto the 3D point cloud to acquire 3D masks, then grouping and merging these 3D masks based on their spatial relationships to generate 3D objects, as shown in Fig~\ref{fig:dino}\textcolor{red}{(c)}. 
However, due to variations in object sizes and corresponding camera poses of 2D frames across different scenes, it is nearly impossible to establish a unified grouping criterion for different objects. If the grouping criterion is too strict, allowing only masks with high overlap to be grouped, masks from multiple parts of a large object cannot be merged (see Fig~\ref{fig:dino}\textcolor{red}{c}-``strict''). Conversely, if the criterion is relaxed, masks that should belong to multiple objects may be merged into a single object, \eg, the ``chair'' and ``sofa'' in Fig~\ref{fig:dino}\textcolor{red}{c}-``relaxed''. We attribute this limitation to the inability to rely solely on 3D spatial relationships and 3D point features (see Fig~\ref{fig:dino}\textcolor{red}{b}) to determine if multiple 3D masks belong to the same object. 
To tackle this issue, we propose a grouping-first-then-projection pipeline: 1) it co-segments objects across multiple 2D frames by exploiting the temporal consistency of the frames, naturally identifying and grouping multiple 2D masks for each single object. 2) It projects the 2D masks corresponding to each object onto the 3D point cloud and then directly merges them.

\noindent\textbf{Object Co-segmentation from Consecutive 2D RGB Frames.} 
Given that 2D RGB frames are captured consecutively, we leverage their temporal consistency to co-segment multi-frame 2D masks corresponding to the same objects.
First, for each RGB frame $I_m$, we apply the 2D unsupervised instance segmentation method MaskCut~\cite{wang2023cut} to obtain the 2D masks of objects in it, denoted as $O_m$. 
As MaskCut uses the image encoder of 2D self-supervised DINO~\cite{caron2021emerging_dino} to extract image feature map, we employ the same encoder to extract features for objects $O_m$ by masked average pooling each object mask $o_m^i \in O_m$ over the frame's feature map. 
Next, for any adjacent pair of frames $I_m$ and $I_{m+1}$, we compute the pairwise similarity between their object masks $O_m$ and $O_{m+1}$ based on mask features. Then, for each object mask $o^i_{m} \in O_{m}$ in frame $I_{m}$, we find the object mask $o_{m+1}^{j^\ast}$ in frame $I_{m+1}$ with the highest similarity as its corresponding candidate mask as follows,
\begin{equation}
\begin{aligned}
    o_{m+1}^{j^\ast} &= \underset{o_{m+1}^j \in O_{m+1}}{\arg\max} \phantom{1}  \text{sim}(o^j_{m+1},o_m^i), \phantom{1} \\ 
    g_{m\xrightarrow{}{m+1}}^{i} &= o_{m+1}^{j^\ast} \phantom{1} \texttt{if} \phantom{1}\text{sim}(o_{m+1}^{j^\ast},o_m^i) > \tau  \phantom{1} \texttt{else} \phantom{1} \textcolor{black}{\text{Null}},
\end{aligned}
\label{equ:d2}
\end{equation}
where $\text{sim}(o^j_{m+1},o_m^i)$ is to compute the cosine similarity between mask features of object masks $o^j_{m+1}$ and $o_m^i$. The $o_{m+1}^{j^\ast}$ in frame $I_{m+1}$ is the candidate object mask of the object mask $o_m^i$ in frame $I_m$. Notably, only when the similarity between the object mask $o_m^i$ and its candidate object mask $o_{m+1}^{j^\ast}$ exceeds the threshold $\pi$, they are considered to belong to the same object. After determining the same objects between any adjacent frames, $g_{m\xrightarrow{}{m+1}}^{i}$, we propagate this ``sameness'' to all adjacent frames to identify the same objects across all frames. Consequently, we can gather objects and identify all 2D masks in different frames of each object.

\noindent\textcolor{black}{\textbf{3D Objectness Priors.}}
Given that we have grouped 2D masks into different objects, for each object, we can directly project its 2D masks onto the 3D point cloud and seamlessly merge their 3D masks into a unified whole~\cite{yang2023sam3d}, forming the object's 3D mask. 
Furthermore, instead of directly utilizing the 3D masks of objects, we utilize the 3D bounding boxes $B^{\text{3D}}$ of their masks to guide hierarchical clustering. This decision is based on treating the collected 3D objects from 2D frames as objectness priors and leaving precise segmentation to the clustering process itself.

\subsection{Hi-Mask3D and Learning Objective}
\label{subsec:mask3d}
In this section, we introduce our 3D instance segmentation framework, Hi-Mask3D. It extends and adapts the 3D instance segmentation model, Mask3D~\cite{Schult23mask3d}, to predict and utilize object parts to improve 3D instance segmentation.

\noindent\textbf{Introduction to Mask3D.}
The 3D instance segmentation model Mask3D~\cite{Schult23mask3d} mainly consists of three components: 1) a feature backbone, a sparse convolutional U-net~\cite{choy20194d} that takes a colored point cloud as input and yields multi-resolution feature matrices $\{\boldsymbol{F}_r\}$, where each $\boldsymbol{F}_r$ $\in \mathbb{R}^{M_r\times C}$ of size $M_r$. 2) A mask module predicts instance masks $B$ corresponding to instance queries $Q \in \mathbb{R}^{K\times C}$ by computing the dot product between queries $Q$ and scene features \{$\boldsymbol{F}_r$\}.
3) A query refinement module, a transformer decoder~\cite{vaswani2017attention} that cross-attends $K$ instance queries $Q$ with multi-resolution feature matrices $\{\boldsymbol{F}_r\}$. 
In the $r$-th decoder layer of query refinement module, the cross-attention between instance queries $Q_r$ and feature matrix $\boldsymbol{F}_r$ is formulated as follows,
\begin{equation}
\begin{aligned}
    & Q_r = \text{softmax}(Q_{r-1}\boldsymbol{F}_{r}^{T}/ \sqrt{C}) \boldsymbol{F}_{r}.
\end{aligned}
\end{equation}
\textcolor{black}{Note that for simplicity of demonstration, we omit the QKV projection~\cite{vaswani2017attention} and intermediate instance mask $B$~\cite{Schult23mask3d} in the cross-attention formulation.}

\noindent\textbf{Hi-Mask3D with Hierarchical-Aware Detector.} Our Hi-Mask3D modifies the query refinement module of Mask3D to adapt for both object and object part segmentation. 
Specifically, we start with two sets of queries $Q^o$ and $Q^p$ representing object queries and object part queries, respectively. Next, in each decoder layer, we no longer directly perform cross-attention between object queries and feature matrix to update object queries. Instead, we perform hierarchical attention: 1) conducting cross-attention between object part queries and feature matrix, and 2) updating object queries using part queries. This enables explicit interaction between object part queries and object queries, which helps to leverage the information from object parts to improve object segmentation. 
In the $r$-th decoder layer, the hierarchical attention is formulated as follows:
\begin{equation}
\begin{aligned}
    &Q^p_{r} = \text{softmax}(Q^p_{r-1}\boldsymbol{F}_r^T / \sqrt{C})\boldsymbol{F}_r,\\
    &Q^o_{r} = \text{softmax}(Q^o_{r-1}(Q^p_{r})^T / \sqrt{C})Q^p_{r}.
\end{aligned}
\end{equation}
The last-layer part queries and object queries undergo the mask module to obtain object part and part segmentation results, $P_{\text{part}}$ and $P_{\text{object}}$, respectively.

\noindent\textbf{Training Objective.}
Following Mask3D~\cite{Schult23mask3d}, we adopt bipartite graph algorithm~\cite{carion2020end, stewart2016end} to match prediction $P_{\text{object}}$ and $P_{\text{part}}$ with pseudo-labels $\hat{P}_{\text{object}}$ and $\hat{P}_{\text{part}}$, respectively. To supervise segmentation masks, we combine DICE~\cite{dice} loss and Binary Cross Entropy (BCE) loss~\cite{lin2017focal}. Similar to \cite{rozenberszki2023unscene3d}, we also employ iterative rounds of self-training to enhance the segmentation capabilities of Hi-Mask3D. During self-training, predictions from the previous round are utilized as pseudo-labels for the subsequent round.

\section{Experiment}
\label{sec:experiment}
\noindent\textbf{Datasets and Implementations.}
We evaluate our method on four public indoor point cloud datasets, including ScanNet~\cite{dai2017scannet}, Scannet200~\cite{rozenberszki2022language}, S3DIS~\cite{s3dis} and Replica~\cite{straub2019replica} .
Following Mask3D~\cite{Schult23mask3d}, we employ a Sparse CNN, Res16UNet34C~\cite{choy20194d}, implemented in MinkowskiEngine~\cite{choy20194d}, as the point cloud encoder. Our Hi-Mask3D is trained for 600 epochs with a learning rate 1e-4 and a batch size 4.
Experiments are conducted using PyTorch~\cite{paszke2019pytorch} and executed on NVIDIA Tesla A40 GPUs.
We evaluate the performance of class-agnostic instance segmentation using standard average precision scores. The scores consist of mAP@25, mAP@50, and mAP, representing performance metrics at IoU thresholds of 0.25, 0.5, and mean average precision, respectively. Hyper-parameter $T = 0.05$ in Equ~\ref{equ:cluster_metric} and $\tau$ = 0.3 in Equ~\ref{equ:d2}.
\subsection{Comparison with State-of-the-Art Unsupervised Methods}
\begin{table}[t]
    \caption{\textbf{Comparison of instance segmentation results on ScanNet val.}} 
    \label{tab:main1}
    \vspace{-3mm}
    \begin{minipage}{0.48\textwidth}
        \centering
        \subcaption{\textbf{Training-free Setting.} All methods do not require gradient backpropagation to obtain prediction results.}
        \vspace{0.5mm}
        \tablestyle{1pt}{1.08}
        \begin{tabular}{lccc}\toprule
\multicolumn{1}{c}{w/o training} &\multicolumn{3}{c}{ScanNet Val~\cite{dai2017scannet}}  \\\cmidrule(lr){2-4} 
Methods &mAP@25 &mAP@50 &mAP\ \ \  \\\midrule
HDBSCAN~\cite{mcinnes2017accelerated_hdbscan}   & 32.1 &5.5 & 1.6\\
Nunes~\cite{nunes2022unsupervised}  & 30.5 & 7.3 & 2.3\\
Felzenswalb~\cite{felzenszwalb2004efficient} & 38.9 & 12.7 & 5.0  \\
CutLER~\cite{wang2023cut}  & 7.0 & 0.2 & 0.3\\
Unscene3D~\cite{rozenberszki2023unscene3d}  & 19.9 & 10.0 & 5.9\\
\rowcolor{gray} Ours  & 55.1 & 26.8 & 12.6\\
\bottomrule
\end{tabular}
        \label{tab:tab1_p1}
    \end{minipage}
    \hspace{1.5mm}
    \centering
    \begin{minipage}{0.48\textwidth}
    \subcaption{\textbf{Data-efficient Setting} with x\% available data annotations. 
    A 0\% value indicates that the method can provide pseudo-labels, and training is conducted on them.}
    \begin{minipage}{0.43\textwidth}
        \vspace{-2mm}
        \centering
        \tablestyle{0pt}{1.08}
        \begin{tabular}{lrrrrr}\toprule
\multicolumn{1}{c}{w/ training} &\multicolumn{5}{c}{mAP@50 ScanNet Val~\cite{dai2017scannet}}  \\\cmidrule(lr){2-6} 
Methods &\phantom{1} 0\% &\phantom{1} 1\% &\phantom{1} 5\% &\phantom{1} 10\% & 20\%  \\\midrule
Scratch~\cite{Schult23mask3d}  & / & 14.1 &33.3 &39.2&43.4 \\
CSC~\cite{hou2021exploring} & / & 22.1 & 39.9 & 43.8&48.9 \\
HDBSCAN~\cite{mcinnes2017accelerated_hdbscan} & 10.5 & 15.1 & 36.3 & 40.0& 42.7 \\
Felzenswalb~\cite{felzenszwalb2004efficient} & 15.2 & 25.3 & 37.2 & 45.7& 50.0 \\
Unscene3D~\cite{rozenberszki2023unscene3d} & 23.2 & 28.4 & 46.8  & 55.7 & 60.7\\
\rowcolor{gray} Ours & 32.6 & 44.1 & 64.2  & 68.0& 72.1 \\
\bottomrule
\end{tabular}
        \label{tab:tab1_p2}
    \end{minipage}
    \end{minipage} 
    \vspace{-5mm}
\end{table}

Table~\ref{tab:main1} shows the results on ScanNet dataset~\cite{dai2017scannet} under two different settings: 

\noindent\textbf{Traning-free Setting.}
In the training-free setting, all methods rely solely on clustering~\cite{caron2021emerging_dino} or graph-cut~\cite{shi2000normalized_cut,wang2023cut} algorithms for prediction, as no annotations are available for use. Table~\ref{tab:tab1_p1} compares our method, Part2Object, with the previous state-of-the-art method, Unscene3D~\cite{rozenberszki2023unscene3d}, which also leverages DINO features but applies graph-cut algorithms on 3D point clouds. Compared to Unscene3D, our approach demonstrates significant enhancements across all metrics, with increases of 35.2\%, 16.8\%, and 6.7\% in mAP@25, mAP@50, and mAP, respectively. These improvements are attributed to our utilization of 3D objectness priors, which guide the clustering process.


\noindent\textbf{Data-efficient Fine-tuning Setting.}
In the data-efficient fine-tuning setting, all methods initially undergo either pretraining or self-training. Subsequently, they fine-tune on downstream data with varying percentages (0\% represents pseudo-labels solely). Under this setting, the extraction of pseudo-labels is the primary requirement, followed by the necessity for models to exhibit strong learning capabilities. As demonstrated in Table~\ref{tab:tab1_p2}, Hi-Mask3D exhibits remarkable performance even with limited data, showcasing the robust 3D representation capabilities learned on the pseudo-labels from Part2object. Specifically, our method surpasses the previous SOTA by 15.7\% and 11.4\% when considering the utilization of only 1\% and 20\% of the available data, respectively.

\subsection{Comparison on Cross-dataset Generalization}

Table~\ref{tab2_p1} and Table~\ref{tab2_p2} show our comparisons with fully-supervised Mask3D~\cite{Schult23mask3d} on cross-dataset generalization, including both zero-shot and dataset-efficient settings. 
In both settings, our Hi-Mask3D is trained on pseudo-labels from our unsupervised Part2Object on the ScanNet dataset~\cite{dai2017scannet}, while Mask3D is trained fully supervised on ScanNet. 

\noindent\textbf{Cross-dataset Zero-shot Generalization Setting.} 
In Table~\ref{tab2_p1}, we compare unsupervised Hi-Mask3D with fully-supervised Mask3D on three downstream datasets, including ScanNet200~\cite{scannet200}, S3DIS~\cite{s3dis} and Replica~\cite{straub2019replica}. Compared with the fully supervised Mask3D trained on ScanNet, our Hi-Mask3D trained on pseudo-labels from Part2Object achieves consistent performance improvement without using any manually annotated data. 
\textcolor{black}{Hi-Mask3D surpasses fully-supervised Mask3D by 10.7\%, 4.8\%, and 6.4\% in terms of mAP@50 on ScanNet200, S3DIS, and Replica datasets, respectively.}
Thanks to the learning from pseudo-labels in Part2Object, Hi-Mask3D has acquired more generalizable representations of 3D objects. Unlike fully-supervised methods, which are constrained to annotated classes in the training dataset, Hi-Mask3D demonstrates robust performance in cross-dataset generalization. 

\noindent\textbf{Cross-dataset Data-efficient Generalization Setting.} 
In Table~\ref{tab2_p2}, we conduct data-efficient training on downstream datasets, such as ScanNet200~\cite{scannet200} and S3DIS~\cite{s3dis}, after pre-training our Hi-Mask3D on ScanNet's pseudo-labels extracted from Part2Object. 
For ScanNet200 and S3DIS, Hi-Mask3D outperforms Mask3D 47.4\% and 24.7\% mAP@50 with 20\% data, respectively. 
When limited data is available, the improvement brought by pretraining on pseudo-labels from Part2Object is even greater, with the enhancement increasing from 37.6\% to 57.3\% as the available data decreases from 20\% to 1\%.

\begin{table}[t]
    \caption{\textbf{Comparison on cross-dataset zero-shot generalization setting.} We compare the zero-shot generalization ability of unsupervised Hi-Mask3D with fully-supervised Mask3D on three downstream datasets: ScanNet200, S3DIS and Replica.}
        \centering
        \tablestyle{2pt}{1.08}




\begin{tabular}{l|ccc|ccc|ccc}\toprule
Zero-shot &\multicolumn{3}{c}{ScanNet200 Val~\cite{scannet200}} &\multicolumn{3}{c}{S3DIS 6-fold ~\cite{s3dis}} &\multicolumn{3}{c}{Replica ~\cite{straub2019replica}} \\
\cmidrule(lr){2-4} \cmidrule(lr){5-7} \cmidrule(lr){8-10}
Methods &  m@25 & m@50 &mAP &  m@25 & m@50 &mAP &  m@25 & m@50 &mAP\\
\midrule
 Mask3D~\cite{Schult23mask3d} 
 & $30.8$ & $24.2$ & $15.4$
 &$17.6$ &$11.7$ &$7.7$
 &$20.8$ &$15.6$ &$9.7$\\
 
\rowcolor{gray}  Ours 
& \textbf{$63.2$} & \textbf{$34.9$} & \textbf{$16.3$} 
& \textbf{$24.5$} & \textbf{$16.5$} & \textbf{$8.5$} 
& \textbf{$36.5$} & \textbf{$22.0$} & \textbf{$11.2$} \\
\bottomrule
\end{tabular}

        \label{tab2_p1}
\end{table}

\begin{table}[t]
    \caption{\textcolor{black}{\textbf{Comparison on cross-dataset data-efficient generalization setting.}} We utilize pseudo-labels from Part2Object to pre-train our Hi-Mask3D and conduct data-efficient finetuning on downstream datasets, comparing them with Mask3D. }
        \centering
        \tablestyle{3pt}{1.08}
        \begin{tabular}{lccccc|cccc}
\toprule
\multirow{2}{*}{Methood} & \multirow{2}{*}{Dataset} &  \multicolumn{4}{c}{mAP@25} 
&  \multicolumn{4}{c}{mAP@50} 
\\ 

\cmidrule{3-6} \cmidrule{7-10}
& & 1\% & 5\% & 10\% & 20\%  & 1\% & 5\% & 10\% & 20\%\\
\midrule

        Scratch  & ScanNet200~\cite{scannet200}  
        & $2.1$ & $22.5$ & $30.6$ & $40.2$ 
        & $1.3$ & $8.2$ & $18.4$ & $22.9$ \\
        \rowcolor{gray} Ours & ScanNet200~\cite{scannet200} 
        & $59.4$ & $70.4$ & $72.3$ & $77.8$ 
        & $35.4$ & $52.8$ & $56.2$ & $62.8$ \\
\midrule
        Scratch & S3DIS~\cite{s3dis} 
        & $1.7$ & $10.1$ & $20.8$ & $45.4$ 
        & $1.0$ & $2.9$ & $9.1$ & $25.9$\\
        \rowcolor{gray} Ours & S3DIS~\cite{s3dis} 
        & $49.1$ & $55.9$ & $65.1$ & $70.2$ 
        & $25.7$ & $35.0$ & $46.4$ & $49.4$\\

\bottomrule
\end{tabular}
        \label{tab2_p2}
        \vspace{-5mm}
\end{table}

\subsection{Ablation Study}
\begin{table}[t]
    \caption{\textbf{Ablation study.} We conduct ablation studies on ScanNet val. We validate the effectiveness of our Part2Object clustering, Hi-Mask3D, and self-training procedure.}
    \label{tab:ablation}
    \vspace{-2mm}
    \begin{minipage}{0.48\textwidth}
        \vspace{-2.5mm}
        \centering
        \subcaption{\textbf{Ablation study on Part2Object clustering and Hi-Mask3D architecture.} \texttt{OG}, \texttt{FU} denotes objectness guidance and cluster feature function, respectively. }
        \vspace{-2mm}
        \tablestyle{1.5pt}{1.0}
        \begin{tabular}{lccc}
        \toprule
         \multicolumn{1}{l}{Ablation on } &\multicolumn{3}{c}{ScanNet Val~\cite{dai2017scannet}}  \\\cmidrule(lr){2-4} 
            Clustering &m@25 &m@50 &mAP\ \ \  \\\midrule
        Baseline   & 37.8 & 13.8 & 6.1 \\
         w/o \texttt{OG}  & 43.2 & 20.7 & 10.4 \\
         w/o \texttt{FU}  & 44.2 & 21.5  & 10.6\\
        \rowcolor{gray}Ours  & \textbf{55.1} & \textbf{26.8} & \textbf{12.6} \\
        \midrule
         Architecture & & &\ \ \  \\\midrule
        Mask3D  & 59.0 & 31.0  & 14.2\\
        \rowcolor{gray}Hi-Mask3D   & \textbf{64.9} & \textbf{36.0} & \textbf{16.9}\\
        \bottomrule
    \end{tabular} 
        \label{tab:a1}
    \end{minipage}
    \hspace{2mm}
    \centering
    \begin{minipage}{0.48\textwidth}
    \subcaption{\textbf{Ablation study on Self-training and Hyper-parameter.} }\label{subtab:zero}
    \begin{minipage}{0.43\textwidth}
        \vspace{-2mm}
        \centering
        \tablestyle{1.5pt}{1.0}
        \begin{tabular}{lccc}
        \toprule
         \multicolumn{1}{l}{Ablation on } &\multicolumn{3}{c}{ScanNet Val~\cite{dai2017scannet}}  \\\cmidrule(lr){2-4} 
            Self-Training &m@25 &m@50 &mAP\ \ \  \\\midrule
        Pseudo labels  & 55.1 & 26.8 & 12.6 \\
        Round $1$  & 60.7 & 32.6 & 15.3 \\
        Round $2$ & 64.3 & 35.2 & 16.5 \\
       \rowcolor{gray} Round $3$ & \textbf{64.9} & \textbf{36.0} & \textbf{16.9} \\ \midrule
        Hyper-parameter & & &\ \ \  \\\midrule
        K = 0.4  & 55.7 & 22.6  & 10.3\\ 
        K = 0.5  & \textbf{57.0} & 25.3  & 12.0\\
        \rowcolor{gray}K = 0.6   & 55.1 & \textbf{26.8} & \textbf{12.6}\\
        K = 0.7  & 44.5 & 21.1  & 10.6\\ 
        \bottomrule
    \end{tabular} 
        \label{tab:a2}
    \end{minipage}
    \end{minipage} 
    \vspace{-3mm}
\end{table}

To evaluate the effectiveness of our hierarchical clustering Part2Object and 3D instance segmentation architecture Hi-Mask3D, as well as the effect of hyperparameters, we conduct comprehensive and detailed ablation experiments, as outlined in Table~\ref{tab:a1} and Table~\ref{tab:a2}. 

\noindent \textbf{Ablation on Part2Object Clustering.} 
Table~\ref{tab:a1} illustrates the variants of Part2Object clustering, starting with a baseline utilizing the single-layer clustering algorithm. The performance of this baseline yields only 6.1\% in mAP and 13.8\% in mAP@50. This limitation arises due to the single-layer clustering being either too loose or too tight, leading to over-clustered or under-clustered objects. 
Subsequently, we remove the guidance of objectness priors from our proposed clustering algorithm, denoted as ``w/o \texttt{OG}''.
Without object priors, while objects may appear at every layer, it becomes challenging to determine which layer they belong to.
Furthermore, we replace our cluster feature computation function, $``\texttt{FU}(\cdot)$'', as simple averaged over features of points within the cluster, ``w/o $\texttt{FU}$''. The 2.0\% decline in mAP highlights the importance of removing noise features for more robust cluster feature representations. 

\noindent \textbf{Ablation on Hi-Mask3D.} 
We compare our Hi-Mask3D with Mask3D using the same pseudo-labels extracted from Part2Object. Hi-Mask3D improves Mask3D by 2.7\% and 5.0\%  in terms of mAP and mAP@50, respectively. This demonstrates that additional part information aids in object understanding. Furthermore, through reporting the performance of self-training, we observe that after several rounds, the mAP increases from 15.3\%  to 16.9\%.

\noindent \textbf{Ablation on Hyper-parameter.} 
We conduct ablation on the import hyper-parameter $K$ in Equ~\ref{equ:cluster_metric}. Since the number of clusters at each layer varies, we use percentages (K=$0.6$ indicates the top 60\%) to measure how many of the top clusters from each layer can be aggregated. Table~\ref{tab:a2} shows the performance of Part2Object at different $K$. 
We use grey to indicate the default setting.

\section{Conclusion and Limitations}
\label{sec:conclusion}
We introduce Part2Object, an efficient hierarchical clustering algorithm, that progressively groups point clouds into object parts and objects, while leveraging 3D object-ness prior to precisely target objects.
Furthermore, we propose the Hierarchical-Aware Mask3D, facilitating self-training with pseudo-object and part labels from Part2Object. Experimental results demonstrate that we consistently surpass all existing methods in both unsupervised settings and data-efficient settings.
\noindent \textbf{Ethics Statement:} Given that our 2D knowledge is derived from the self-supervised models DINO, we acknowledge that biases and controversies inherent in the training data for these models may be introduced into our model.\noindent\textbf{Acknowledgment:} This work was supported by the National Natural Science Foundation of China (No.62206174) 
and MoE Key Laboratory of Intelligent Perception and Human-Machine Collaboration (ShanghaiTech University).


%
%
\bibliographystyle{splncs04}
\bibliography{main}
\end{document}